%% file: arxiv.tex
\documentclass{article}

\usepackage[preprint]{neurips_2025}
\usepackage[pagebackref,breaklinks,colorlinks,citecolor=cyan,linkcolor=cyan]{hyperref}

\usepackage{float} %
\usepackage{amsmath}
\usepackage{graphicx}
\usepackage{subcaption}
\usepackage{subcaption}
\usepackage{algorithm}
\usepackage{algorithmic}
\usepackage{listings}
\usepackage{xcolor}
\usepackage{enumitem}  %

\definecolor{codegray}{gray}{0.95}
\usepackage{multirow}
\usepackage[utf8]{inputenc} %
\usepackage[T1]{fontenc}    %
\usepackage{hyperref}       %
\usepackage{url}            %
\usepackage{booktabs}       %
\usepackage{amsfonts}       %
\usepackage{nicefrac}       %
\usepackage{microtype}      %
\usepackage{xcolor}         %
\usepackage{pifont}
\definecolor{customred}{HTML}{C59D94}   %
\definecolor{customgreen}{HTML}{458991} %
\definecolor{Magenta}{rgb}{0.8, 0.1, 0.6}

\title{Guiding the Experts: Semantic Priors for Efficient and Focused MoE Routing }

\usepackage{authblk}

\author{
Chengxi Min$^{1}$,
Wei Wang$^{1}$\thanks{Corresponding author. Email: \texttt{wei.wang@bjtu.edu.cn}},
Yahui Liu$^2$,
Weixin Ye$^1$,
Enver Sangineto$^3$,
Qi Wang$^2$,
Yao Zhao$^1$\\
$^1$Department of Computer Science, Beijing Jiaotong University \\
$^2$Kuaishou Technology\\
$^3$AImageLab, Dipartimento di Ingegneria ”Enzo Ferrari”, University of Modena and Reggio Emilia.\\

}

\begin{document}

\maketitle
\vspace{-1.2em} 
\input{sections/0-abstract}

\input{sections/1-introduction}

\input{sections/2-related-work}

\input{sections/3-preliminary}
\input{sections/4-method}
\input{sections/5-experiments}

\input{sections/5-2-analysis}

\input{sections/6-conclusion}

\bibliography{refs}
\bibliographystyle{plain} %

\input{sections/appendix}

\clearpage

\end{document}

%% file: sections/0-abstract.tex
\begin{abstract}

Mixture-of-Experts (MoE) models have emerged as a promising direction for scaling vision architectures efficiently.  Among them, Soft MoE improves training stability by assigning each token to all experts via continuous dispatch weights. However, current designs overlook the semantic structure which is implicitly encoded in these weights, resulting in suboptimal expert routing. In this paper, we discover that dispatch weights in Soft MoE inherently exhibit segmentation-like patterns but are not explicitly aligned with semantic regions. Motivated by this observation, we propose a foreground-guided enhancement strategy. 
Specifically, we introduce a spatially aware auxiliary loss that encourages expert activation to align with semantic foreground regions. To further reinforce this supervision, we integrate a lightweight LayerScale mechanism that improves information flow and stabilizes optimization in skip connections. 
Our method necessitates only minor architectural adjustments and can be seamlessly integrated into prevailing Soft MoE frameworks. Comprehensive experiments on ImageNet-1K and multiple smaller-scale classification benchmarks not only showcase consistent performance enhancements but also reveal more interpretable expert routing mechanisms.
The source codes and models are avialable at: \url{https://github.com/0930mcx/Guiding-Experts}.
\end{abstract}

%% file: sections/1-introduction.tex
\section{Introduction}
\label{sec:introduction}

Scaling up model size has proven to be an effective approach for improving performance in deep learning~\cite{krizhevsky2012imagenet,kaplan2020scaling,zhai2022scaling}. However, this comes at the cost of significantly increased computational resources during both training and inference phrases. To address this, Sparse MoE models~\cite{riquelme2021scaling} have been proposed to dynamically activate only a subset of expert modules based on input tokens. This design allows for greater model capacity without a corresponding increase in computational cost.
Despite various routing strategies~\cite{muqeeth2023soft,wu2024mixture,liu2024deepseekv2,liu2024deepseekv3} having been developed to optimize the assignment between tokens and experts, the inherently discrete nature of this process in Sparse MoE introduces challenges such as unstable training and difficult convergence. To alleviate these issues, Soft MoE~\cite{puigcerver2023sparse}, a ViT~\cite{dosovitskiy2020image} variant, 
introduces a soft routing mechanism. Different from hard assignments, each token is assigned to a set of continuous dispatch weights, and experts receive a weighted combination of all tokens. This approach significantly improves training stability compared to sparse expert models.

Dispatch weights are a critical component of the Soft MoE routing mechanism, as they represent the correspondence between input tokens and experts. In the original work~\cite{puigcerver2023sparse}, the authors analyzed the dispatch weights of all tokens from the test images to investigate the routing behavior. However, a more fine-grained analysis of the dispatch weights was missing, which leads to the oversight of potential semantic structures encoded in these dispatch weights. Prior studies~\cite{caron2021emerging,oquab2023dinov2,wang2023tokencut,peruzzo2024spatial} have shown that the attention maps in ViT exhibit patterns resembling semantic segmentation during self-supervised training. %
We empirically observed that the dispatch weights in Soft MoE computed based on token representations after multi-head self-attention (MSA) also preserve such spatial coherent semantic patterns.

However, the impact of these implicit segmentation-like structures on the performance and training behavior of Soft MoE remains unclear. In this work, 
we  conduct a systematic exploration of how the semantic structure embedded within dispatch weights influences model training.
Subsequently, we propose guided mechanisms designed to harness this characteristic, thereby optimizing routing processes and enhancing overall model performance.

Notably, we propose a foreground-guided enhancement strategy for Soft MoE. Specifically, 
We devise a spatially guided auxiliary loss function. 
This function encourages the dispatch weights of the routing mechanism to conform to the foreground regions detected based on semantic priors by an external network.
To enhance the effectiveness of supervision and ensure it has a profound impact on expert routing within deeper layers, we propose a LayerScale~\cite{touvron2021going} mechanism. 
This mechanism adaptively regulates the information flow in the skip connections, optimizing the overall routing process and model performance.
Collectively, these elements steer expert activation towards semantically significant regions, thereby enhancing both the stability and efficacy of routing within Soft MoE frameworks.
Moreover, 
our method necessitates only minor architectural adjustments and can be seamlessly integrated into prevailing Soft MoE frameworks.

Our main contributions are as follows:
\begin{itemize}[leftmargin=*, itemsep=0pt, topsep=2pt]
\item We demonstrate that  combining the token-expert soft assignments with a semantic segmentation prior leads to a higher routing efficiency.
\item We propose a foreground-guided auxiliary loss that encourages the dispatch weights in Soft MoEs to focus on semantically meaningful regions, improving expert routing quality.
\item We introduce a lightweight LayerScale mechanism to modulate the residual connections, enhancing supervision propagation and training stability.
\end{itemize}

%% file: sections/2-related-work.tex
\section{Related Work}
\label{sec:related_work}
\textbf{Sparsely activated MoE.} Scaling up model size is an effective way to improve performance in deep learning~\cite{kaplan2020scaling}. However, traditional dense models~\cite{dosovitskiy2020image, liu2021swin, liu2022convnet} activate all parameters during each forward pass, resulting in significant computational overhead as the model scales. To address this, MoE~\cite{riquelme2021scaling, fedus2022switch, jiang2024mixtral,wu2024mixture} models have been proposed, which selectively activate a sparse subset of expert modules per input~\cite{lepikhin2020gshard}. This design has enabled MoE models to scale efficiently.
To further improve performance, a variety of routing strategies have been introduced to better allocate input tokens. Early approaches such as Greedy Top-K~\cite{shazeer2017outrageously}select the top-scoring experts for each token based on routing logits. Hash Layer~\cite{roller2021hash} introduces a fixed token-to-expert mapping using hash-based partitioning, providing fast and deterministic assignments. BaseLayer~\cite{lewis2021base} formulates token routing as a linear programming problem, optimizing the assignment globally. In contrast to token-initiated routing, EC-CF2~\cite{zhou2022mixture} introduces a novel expert-initiated mechanism, where experts select tokens instead of the other way around. Soft MoE~\cite{puigcerver2023sparse}, SMEAR~\cite{muqeeth2023soft}, and Lory~\cite{zhong2024lory} all adopt soft routing strategies to improve expert assignment. However, to the best of our knowledge, no previous work in computer vision has explored the spatial semantic patterns of token-to-expert assignments to improve model performance.

\textbf{Semantic Structure in Vision Models.} Previous works have shown that the attention maps learned by Vision Transformers often exhibit implicit semantic segmentation patterns~\cite{naseer2021intriguing, carion2020end}, even under self-supervised training~\cite{caron2021emerging,zhou2021ibot}. Such emergent structures have been utilized in weakly supervised segmentation, attention regularization, and dense prediction tasks~\cite{kirillov2023segment}. Slot Attention~\cite{locatello2020object} also highlights the ability to learn object-centric representations via unsupervised mechanisms. However, these spatial cues have rarely been exploited to inform expert selection in MoE frameworks.

%% file: sections/3-preliminary.tex
\section{Preliminaries}
\label{sec:preliminary}

In this section, we introduce the key concepts necessary for understanding the Aux Loss, including sparse MoE models and the various routing algorithms employed within MoE architectures in the field of computer vision.

In previous MoE models~\cite{riquelme2021scaling,hwang2023tutel,puigcerver2023sparse}, each expert within the MoE module is implemented as an independent multi-layer perceptron (MLP). 
Besides, the MoE module contains a gating network that assigns input tokens to all experts. 
Given a sequential input tokens $\mathbf{x}\in \mathbb{R}^{m\times d}$, a gating network function $G$ and a set of experts $\{E_1, \dots, E_N\}$, where $d$ is the dimensionality of the input tokens, $m$ refers to the number of tokens and $N$ is the number of all experts. We can get the corresponding routing scores for all experts: $\mathbf{s} = G(\mathbf{x}) \in \mathbb{R}^{m\times N}$. Then, in a standard MoE module, its output can be written as follows: 
\begin{equation}
\label{eq:standard-moe}
    \mathbf{y} = \sum_{i}^{N}\mathbf{s}_iE_i(\mathbf{x}).
\end{equation}
However, considering efficiency, we usually consider introducing routing strategies that allow only a portion of the experts to participate in the weighted summation calculation in Eq.~(\ref{eq:standard-moe}). 
Notably, routing algorithms exhibit substantial variations across diverse MoE designs and their functionality is intricately intertwined with model performance. 

Among numerous MoE routing strategies, Top-K routing~\cite{shazeer2017outrageously}, as a simple and efficient strategy, is still widely adopted.
The Top-K routing strategy chooses the top-K experts ranked by all routing scores: $\mathcal{S}_{k} = KeepTopK(\mathbf{s}, k)$ and adjust the final weighted summation:
$\mathbf{y}' = \sum_{\mathbf{s}_j \in \mathcal{S}_k} \mathbf{s}_jE_j(\mathbf{x})$.
Although Top-K routing has been extensively employed in MoE models due to its computational efficiency, it entails a form of hard expert selection. 
This hard selection mechanism can give rise to gradient sparsity issues and lead to instability during the training process.

To overcome these limitations, Soft MoE~\cite{puigcerver2023sparse} replaces discrete expert selection with continuous expert weighting, enabling smoother optimization. 
The SoftMoE module is composed of $N$ experts, denoted as $\{E_1, \dots, E_N\}$, where each expert is equipped with $p$ slots. The learnable matrix $\mathbf{\Phi} \in \mathbb{R}^{d\times (N\times p)}$ parameterizes these slots. Given the input tokens and the slot parameters defined by $\mathbf{\Phi}$, the dispatch weights  $\mathbf{D}\in \mathbb{R}^{m\times(N\times p)}$ are formulated as:
\begin{equation}
\label{eq:Dij}
\mathbf{D}_{i,j}=\frac{\exp(({\mathbf{x}\mathbf{\Phi}})_{i,j})}{\sum_{k=1}^{m}\exp((\mathbf{x}\mathbf{\Phi})_{k,j})},
\end{equation}
where $i \in \{1,,\dots m\}$ and $j \in \{1,\dots,N\times p\}.$
The input of experts module is:
$\widetilde{\mathbf{x}}=\mathbf{D}^T\mathbf{x}$,
where $\widetilde{\mathbf{x}}\in \mathbb{R}^{(n\times p)\times d}$ represents the weighted combination of the input tokens based on the dispatch weights.
The output of experts module is $\widetilde{\mathbf{y}}\in \mathbb{R}^{(N\times p)\times d}$, where
$\widetilde{y}_{i,j}=E_i(\widetilde{\mathbf{x}}_{i,j})$, $i \in \{1,\dots, N\}$ and $j \in \{1,\dots, p\}.$
Then, the output tokens $\mathbf{r} = \mathbf{C}\widetilde{\mathbf{y}} \in \mathbb{R}^{m\times d}$ are obtained by transforming $\widetilde{\mathbf{y}}$ using the Combine Weights $\mathbf{C}\in \mathbb{R}^{m\times(N\times p)}$, where
$\mathbf{C}_{i,j}=\frac{\exp(({\mathbf{x}\mathbf{\Phi}})_{i,j})}{\sum_{k=1}^{N\times p}\exp((\mathbf{x}\mathbf{\Phi})_{i,k})}$.
Finally, the output $\mathbf{r}$ is integrated into the input $\mathbf{x}$ via a residual connection to form the output of the SoftMoE module: 
\begin{equation}
\label{eq:y_softmoe}
    \mathbf{y}' = \mathbf{r} + \mathbf{x}.
\end{equation}

%% file: sections/4-method.tex
\section{Method}
\label{sec:method}

\begin{figure}  
    \centering
    \includegraphics[width=\textwidth]{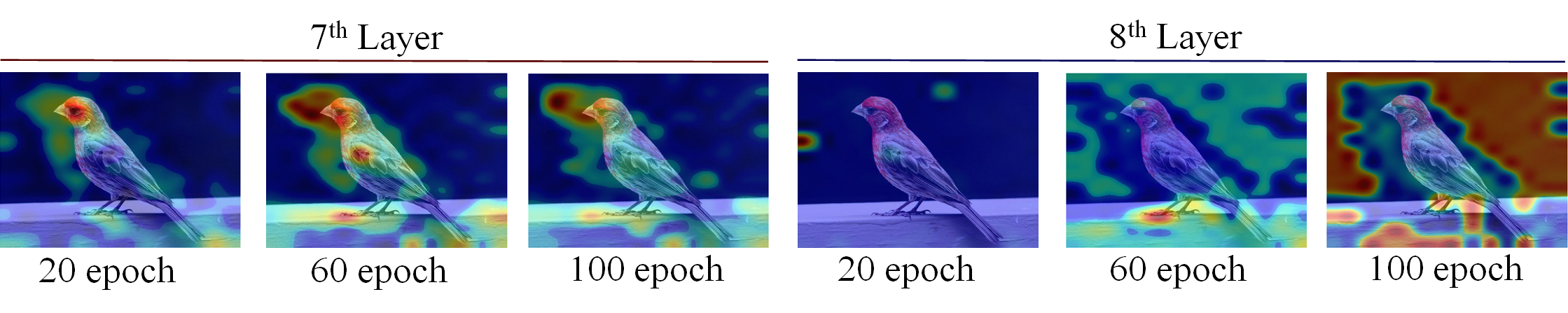} 
    \caption{Visualization of dispatch weight maps during training from scratch at different epochs (20, 60, and 100). The color spectrum ranges from blue to yellow, indicating increasing dispatch weights. The maps illustrate how the expert routing evolves as training progresses over 100 epochs.} \label{fig:dispatch_weight_in_training}
\end{figure}
In natural images, the foreground typically corresponds to the main object category of the entire scene. As shown in Figure~\ref{fig:dispatch_weight_in_training}, 
The bird serves as the primary foreground element and acts as the object of classification.
The dispatch weights within the Soft MoE layer represent the significance of various patches in the image, indicating how crucial each patch is for the overall processing and decision-making within the layer. 
Through monitoring the evolution of dispatch weights during the training process, it can be discerned that the aggregate weights of the foreground region exhibit a coherent and consistent pattern of change, either increasing or decreasing.

\begin{figure}[ht]
    \centering
\includegraphics[width=\linewidth]{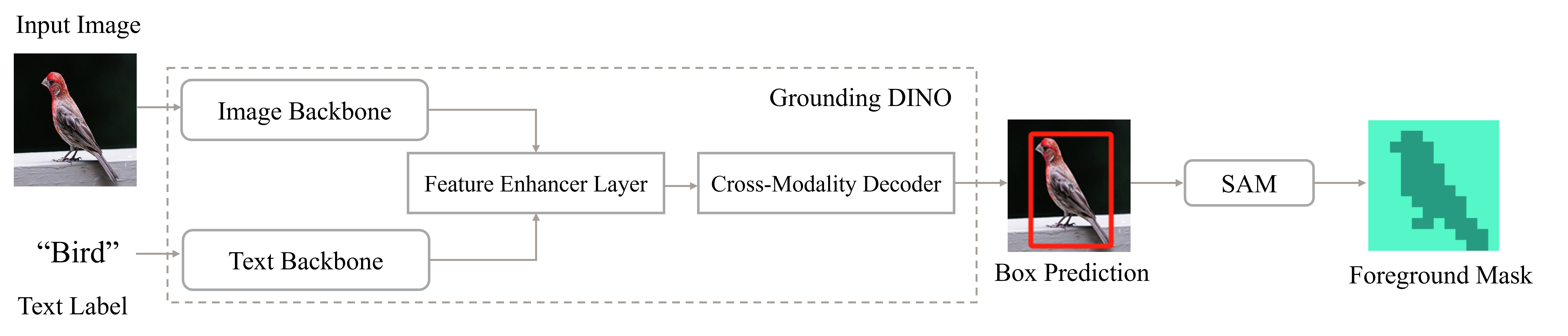}
    \caption{The process for generating the foreground mask involves leveraging Grounding DINO in conjunction with SAM.}
    \label{fig:mask_generator}
\end{figure}

This phenomenon suggests that the routing mechanism of the Soft MoE blocks can spontaneously adapt to the inductive bias inherent in natural images  during the training phase. 
This adaptation implies that the foreground region holds a higher  semantic significance, as the mechanism intuitively assigns higher weights to it, reflecting its crucial role in conveying meaningful information.
Inspired by this, our goal is to incorporate the prior knowledge of the foreground region into the Soft MoE through soft constraints, thereby explicitly increasing or decreasing the dispatch weights corresponding to the foreground region.

\paragraph{Extracting Foreground Masks.}
To extract the foreground regions of the images in the dataset, we use a foreground mask generation pipeline based on Grounding DINO~\cite{liu2024grounding} and SAM~\cite{kirillov2023segment}. 
The process of foreground mask generation is illustrated in Figure~\ref{fig:mask_generator}. 
Given an input image and its corresponding classification label, we first use the text backbone of Grounding DINO to encode the text label as text features, and we use the image backbone to encode the image as visual features. Subsequently, the text features and image features are fed together to the Feature Enhancer of Grouding DINO. Then, GLIP~\cite{li2022grounded} employs a set of image-to-text and text-to-image cross-attention~\cite{vaswani2017attention} mechanisms to facilitate the feature fusion between images and text.
Using the fused features, the cross-modal decoder predicts a set of bounding boxes corresponding to the prompt. 
Then, a bounding box with high confidence is selected, and both this selected bounding box and the original image are fed into the Segment Anything Model (SAM).
Finally, SAM generates fine-grained pixel-level segmentation masks $\mathcal{M} \in \mathbb{R}^m$ based on the input bounding box, where $m$ refers to the pixel number per mask. The implementation of this mask generation pipeline follows an open-source \href{https://github.com/luca-medeiros/lang-segment-anything}{repository}~\cite{medeiros2023langsam}.

\begin{figure}[ht]
\centering
\includegraphics[width=0.9\textwidth]{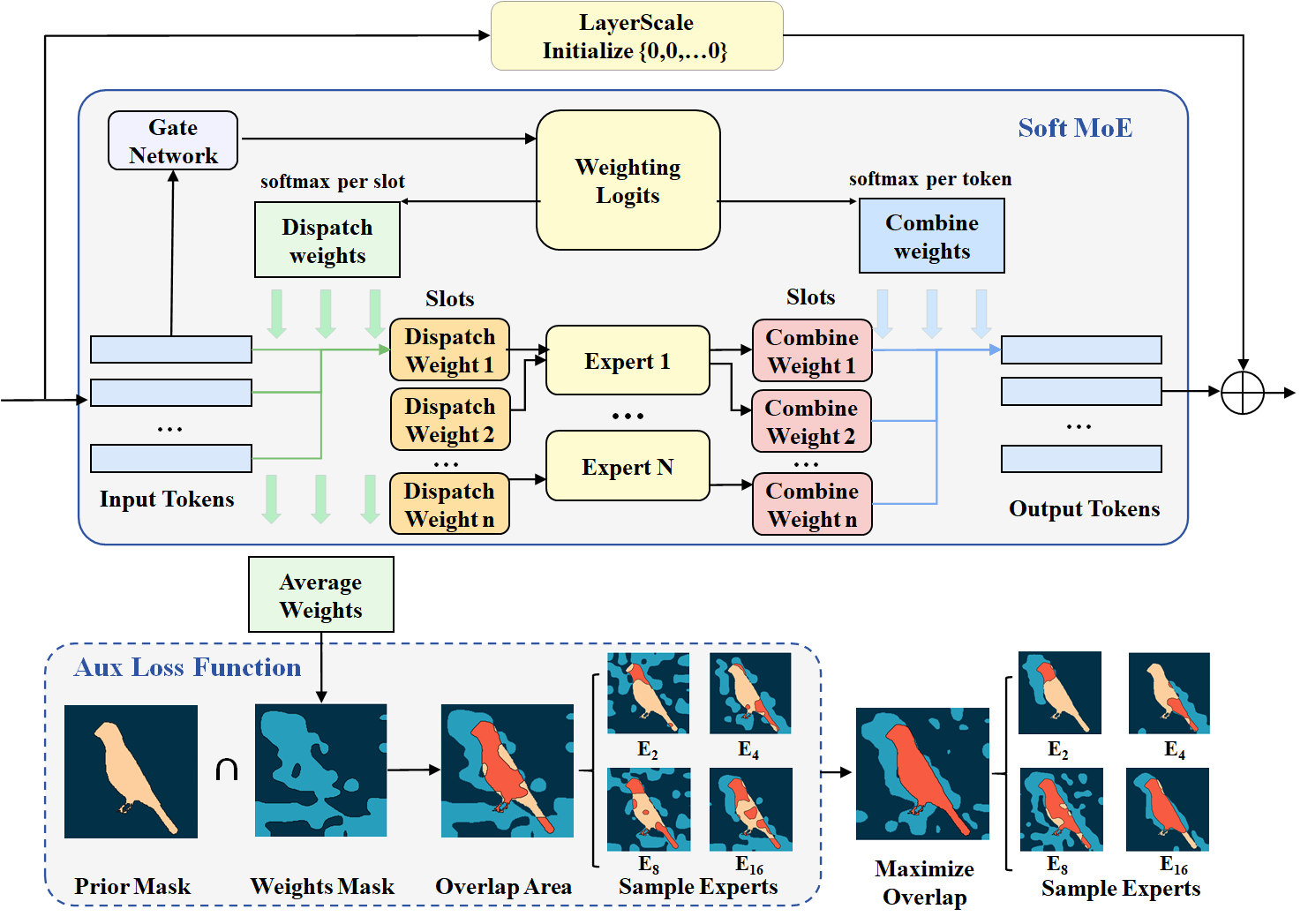} 
\caption{\textbf{Overview of our proposed method.} We first compute the average dispatch weights from the Soft MoE module and apply thresholding based on their mean value to generate a binary weight mask. We encourage this weight mask to overlap with the prior foreground mask as much as possible, guiding expert attention toward semantically meaningful regions in the image. Additionally, we introduce a LayerScale module with an initial value of zero, which adaptively regulates the information flow in skip connections during training. As shown at the bottom of the figure, maximizing the overlap between the dispatch masks and the prior masks leads to more diverse and improved specialization among the four selected experts (chosen from a pool of 32 experts).} 
\label{fig:dispatch_weight_in_training} 
\end{figure}

\paragraph{Auxiliary Loss.}
Given the foreground priors, 
we introduce an auxiliary objective function to guide the MoE routing network, enabling it to perceive spatial semantics and make more reasonable routing predictions.
In the dispatch weight $\mathbf{D}_{i,j}$ (see Eq.~(\ref{eq:Dij}) in Section~\ref{sec:preliminary}) of the Soft MoE, we compute the average dispatch weight $\mathbf{W}\in\mathbb{R}^m$ of every patch for each expert and slot: %
\begin{equation}
      \mathbf{W}_i = \frac{1}{N\times{p}}\sum_{j=1}^{N\times p}\mathbf{D}_{i, j}.
\end{equation}

A higher value $\mathbf{W}_i$ signifies that the corresponding patch token commands more significant attention within the output of the expert layer. 
To further extract spatial saliency features, we first compute the mean value of the weight matrix $\mathbf{W}$: $w = \frac{1}{m}\sum_{i=1}^m \mathbf{W}_i$, which serves as a fundamental step in our feature extraction process to highlight spatial aspects of interest.
Subsequently, we perform a binarization operation on the weight matrix $\mathbf{B}\in \mathbb{R}^m$. Specifically, we set all elements with values smaller than $w$ to 0, and all elements with values greater than or equal to $w$ to 1. 
\begin{equation}
    \mathbf{B}_i = 
\begin{cases}
1, & \text{if } \mathbf{W}_i \geq w \\
0, & \text{otherwise}
\end{cases}.
\end{equation}
Thus, the weight mask $\mathbf{B}$ emphasizes the spatial regions that the dispatch weights deem most significant. 
It functions as a crucial indicator for pinpointing the prominent patches within the image, enabling a more precise analysis of the image's key components.

Next, we compare the weight mask with the foreground mask $\mathcal{M}$. As shwon in Figure~\ref{fig:dispatch_weight_in_training}, we initially calculate the overlapping of the two masks to precisely pinpoint the intersection area.
Subsequently, we compute their union to represent the overall salient region, thereby capturing all the significant parts within the context of our analysis.
The intersection mask and union mask are defined as:
\begin{equation}
    \mathbf{O}_{i} = \mathbf{B}_{i} \cap \mathcal{M}_{i},
\label{eq:overlap}
\end{equation}
\begin{equation}
    \mathbf{U}_{i} = \mathbf{B}_{i} \cup \mathcal{M}_{i}.
\label{eq:union}
\end{equation}
In order to promote the alignment between the weight mask and the foreground mask, our objective is to reduce the disparity between overlapping mask and union mask as much as possible. 
In this way, we can enhance the consistency and effectiveness of these masks in highlighting relevant features within the data.
Intuitively speaking, a substantial overlap suggests that the regions that the model's dispatch weights  have a close correspondence with the actual semantic foreground, indicating that the model's attention mechanism is effectively capturing the meaningful parts of the input.
This close alignment implies that the model is accurately focusing on the regions that carry semantic significance, 
as it helps the model to better understand and process the input data.

Consequently, we calculate the overlap ratio of the two masks in order to precisely quantify the degree of their spatial congruence.
This measurement allows for a more accurate assessment of how well the spatial regions emphasized by the masks align with each other, providing valuable insights into the relationship between the features represented by the masks.
A higher overlap ratio serves as an indication of a more optimal alignment. 
This is highly advantageous, as it effectively ensures that the model predominantly focuses on foreground regions that carry significant semantic value.

Given that the average values of the dispatch weights 
$\mathbf{W}$, we perform an element-wise multiplication of the weight matrix with the overlapping mask $\mathbf{O}$. This operation enables us to obtain the weighted sum within the intersection region. 
Analogously, we carry out the same element-wise multiplication with the union mask $\mathbf{U}$, thereby computing the weighted sum across the entire salient region. 
Therefore, we define the importance alignment score $p$ as the ratio between the two weighted sums:
\begin{equation}  p=\frac{\sum_{i=1}^{{m}}\mathbf{W}_{i}\times \mathbf{O}_{i}}{\sum_{i=1}^{{m}}\mathbf{W}_{i}\times \textbf{U}_{i}}
\end{equation}

Finally, we define the loss as the negative logarithm of the importance alignment score $p$. 
\begin{equation}
    \mathcal{L}_\text{aux} = -\log(p+\epsilon)
\end{equation}
where $\epsilon$ is a small value used to prevent the input of $\log$ from being 0.
The proposed auxiliary loss is combined with the main task loss. For example, in a classification task, the overall loss is defined as:
\begin{equation}\label{lambda}
    \mathcal{L}_{\text{total}} = \mathcal{L}_{\text{cls}} + \lambda \mathcal{L}_{\text{aux}}
\end{equation}
where $\lambda$ is a hyperparameter that balances the two loss components.

\paragraph{LayerScale Mechanism}
In the deeper layers of ViT~\cite{dosovitskiy2020image}, 
the representations of the patch tokens are mainly propagated via skip connections~\cite{he2016deep}. Moreover, the features generated by the deeper blocks rely significantly on those ones generated by the preceding layers. 
This hierarchical propagation mechanism plays a crucial role in aggregating information across different layers of the network, enabling the model to capture both local and global features effectively.
Since Soft MoE is constructed on the ViT backbone, %
it is highly probable that an analogous characteristic is present.
Specifically, the token representations within the deeper layers of Soft MoE are also substantially dependent on residual connections.
This observation is particularly critical for our method, as we perform computations on the dispatch weights of the very last SoftMoE layer. 
These connections play a pivotal role in aggregating and refining semantic information as one traverses through the varying depths of the network.

In a prior study~\cite{peruzzo2024spatial}, the authors calculated the loss by leveraging the attention maps derived from the MSA (Multi-Scale Attention) layer of the final block. 
They noticed that making better use of the skip connections can lead to significantly improved performance. 
Inspired by \cite{peruzzo2024spatial}, we introduce a learnable LayerScale~\cite{touvron2021going} module whose goal is to regulate the skip connection magnitude between the output of the MSA final block  and the Soft MoE block,  
enabling a more precise control of the information flow.
Specifically, we implement the LayerScale operation on the output of the MSA layer. We initialize the value of the LayerScale module to zero, a strategy that effectively renders the residual connection inactive at the onset of the training process. 
This strategy enables the model to gradually incorporate the information carried by the residual path as the training progresses. 
Therefore, the skip connection of the last layer's Soft MoE is formulated as:
\begin{equation}
    \mathbf{y}' = \mathbf{r} + \pmb{\gamma} \mathbf{x}, 
\end{equation}
where $\pmb{\gamma}\in\mathbb{R}^{m\times d}$ refers to the learnable parameters in the LayerScale module that constitutes the primary disparities compared to Eq.~(\ref{eq:y_softmoe}). %

%% file: sections/5-experiments.tex
\section{Experiments}
\label{section:Experiments}

We conduct our experiments on 6 image classification datasets, including ImageNet-1K~\cite{russakovsky2015imagenet}, ImageNet-100~\cite{tian2020contrastive}, Stanford Cars~\cite{krause20133d}, and three datasets in DomainNet~\cite{peng2019moment} (\textit{i.e.}, Clipart, Painting and Sketch). We validate of our method by leveraging the Soft MoE model.
More precisely, our model is composed of eight ViT~\cite{dosovitskiy2020image} blocks. We refer to Appendix~\ref{section:Experimental_Setups} for more details.

\begin{table}[ht]
  \caption{Comparison of baseline Soft MoE~\cite{puigcerver2023sparse} and our proposed method on ImageNet-1K and ImageNet-100. Each method is reported with the corresponding number of training/test samples and top-1/top-5 accuracy. Our method consistently outperforms the baseline on the two datasets.}
  \label{table1}
  \centering
  \begin{tabular}{l c c c cc cc}
\toprule
\multirow{2}{*}{Dataset} & Train  & Test  & \multirow{2}{*}{Classes} 
& \multicolumn{2}{c }{Soft MoE} & \multicolumn{2}{c}{Soft MoE + Ours} \\
\cmidrule(r){5-6} \cmidrule(l){7-8}
& Samples & Samples & & Top-1$\uparrow$ & Top-5$\uparrow$ & Top-1$\uparrow$ & Top-5$\uparrow$ \\
\midrule
ImageNet-1K   & 1.28M  & 50K   & 1000 & 73.9 & 91.5 & \textbf{74.5} & \textbf{91.8} \\
ImageNet-100  & 130K   & 5K    & 100  & 75.4 & 92.4 & \textbf{76.8} & \textbf{92.8} \\
\bottomrule
\end{tabular}
\end{table}

\subsection{Main Results}
\label{section:Main_Results}

\begin{figure}[ht]
  \centering
  \begin{subfigure}[t]{0.17\textwidth}
    \includegraphics[width=\textwidth]{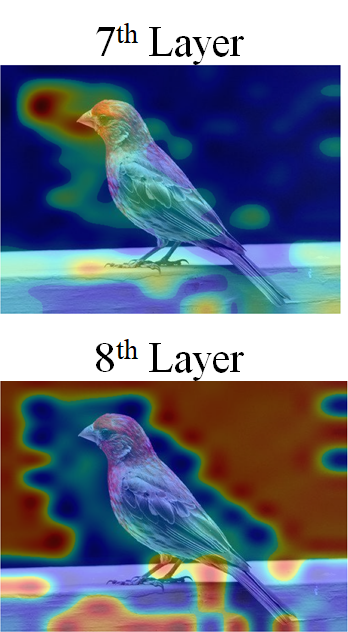}
    \caption{Baseline}
  \end{subfigure}
  \hspace{0.01\textwidth}
  \begin{subfigure}[t]{0.17\textwidth}
    \includegraphics[width=\textwidth]{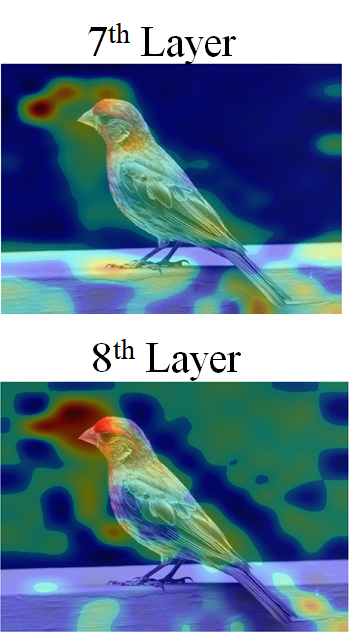}
    \caption{Aux Loss only}
  \end{subfigure}
  \hspace{0.01\textwidth}
  \begin{subfigure}[t]{0.17\textwidth}
    \includegraphics[width=\textwidth]{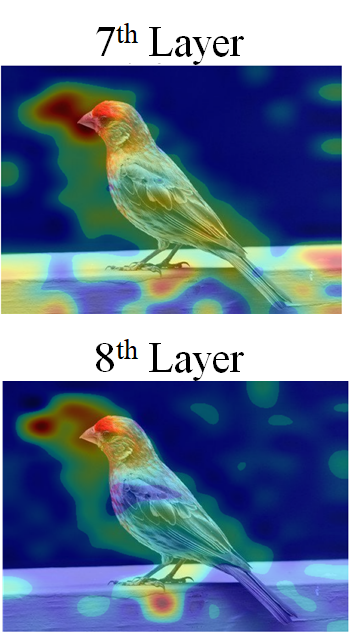}
    \caption{Full method @ 8th layer}
    \label{fig:7layers}
  \end{subfigure}
    \hspace{0.01\textwidth}
  \begin{subfigure}[t]{0.17\textwidth}
    \includegraphics[width=\textwidth]{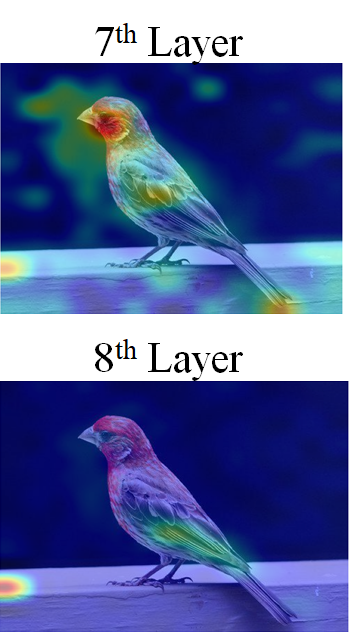}
    \caption{Full method @ 7th layer}
  \end{subfigure}
  \hspace{0.01\textwidth}
  \begin{subfigure}[t]{0.17\textwidth}
    \includegraphics[width=\textwidth]{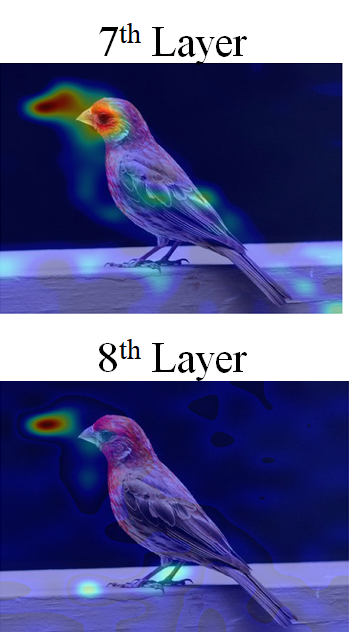}
    \caption{Full method @ 7\&8th layer}
     \label{fig:2layers}
  \end{subfigure}
  \caption{Visualization of dispatch weight maps under different ablation settings. Colors range from blue (low weights) to yellow (high weights), indicating expert assignment intensity.
(a) Baseline model without auxiliary loss or LayerScale;
(b) Model with only auxiliary loss, without LayerScale;
(c) Our full method with auxiliary loss and LayerScale applied at the 8th Soft MoE layer;
(d) Our full method with auxiliary loss and LayerScale applied at the 7th Soft MoE layer;
(e) Our full method with auxiliary loss and LayerScale applied at both the 7th and 8th Soft MoE layer. %
}
  \label{fig:dispatch_ablation_visual}
\end{figure}

\textbf{Train from scratch.}
To comprehensively validate the effectiveness of our method, we conduct pretraining experiments on both ImageNet-1K and ImageNet-100 by training from scratch with the original Soft MoE and our proposed method for 100 epochs. 
As shown in Table~\ref{table1}, 
our proposed method has been proven to reliably enhance the performance of the Soft MoE model across the tested datasets. 
Specifically, it achieves a notable 0.6\% increase in accuracy on the challenging ImageNet-1K dataset, and an even more substantial 1.4\% boost on ImageNet-100. As shown in Figure~\ref{fig:dispatch_ablation_visual}(c), with our method, the expert modules tend to focus more consistently on the foreground regions of the images.
These results underscore the effectiveness of our method in medium-to-large scale training scenarios.

\begin{table}[ht]
\vspace{-1em}
\caption{Fine-tuning results on small-scale datasets. Both baseline and our pretrained model are pre-trained on ImageNet-1K for 100 epochs.}
\label{table2}
\centering
\begin{tabular}{lccc}
\toprule
Dataset & Baseline FT & Baseline + Our FT & Ours Pretrained + Our FT \\
\midrule
Stanford Cars & 35.9 & 37.8 \small{(+1.9)} & \textbf{38.0} \small{(+2.1)} \\
Clipart       & 63.6 & 65.2 \small{(+1.6)} & \textbf{66.4} \small{(+2.8)}  \\
Painting      & 63.7 & 64.2 \small{(+0.5)} & 
\textbf{65.3} \small{(+1.6)}  \\
Sketch        & 56.4 & 57.6 \small{(+1.2)} & \textbf{58.8} \small{(+2.4)}  \\
\bottomrule
\end{tabular}
\vspace{-1em}
\end{table}

\textbf{Finetune.} 
We further explore the effectiveness of our method in fine-tuning settings, using smaller datasets such as Stanford Cars, Clipart, Painting, and Sketch. Table~\ref{table1} shows the performances of Soft MoE (baseline) and our proposed method finetuned on these datasets. 
Specifically, we compare three fine-tuning settings: (1) using the baseline Soft MoE model pre-trained on ImageNet-1K for 100 epochs and fine-tuned with the standard protocol; (2) using the same baseline model but employing our proposed auxiliary loss (\textit{i.e.}, Eq.~(\ref{lambda})) only in the fine-tuning phase; and (3) using our pre-trained Soft MoE variant and utilizing auxiliary loss during the fine-tuning for 100 epochs.

As shown in Table~\ref{table2}, our method consistently improves performance across all datasets. 
Firstly, we can observe that the utilization of the additional constraints $\mathcal{L}_\text{aux}$ proposed by us solely during the fine-tuning stage can also yield improvements.
Then, by leveraging our pre-trained model and incorporating the training constraint $\mathcal{L}_\text{aux}$ we proposed, a more significant performance improvement can be achieved. 
This demonstrates not only the effectiveness of our method in fine-tuning, but also the robustness of the model pre-trained with our approach when transferred to smaller datasets.

\begin{minipage}[t]{0.48\textwidth}
  \centering
  \captionof{table}{Ablation on auxiliary loss and LayerScale mechanism.}
  \label{tab:ablation_components_of_method}
  \begin{tabular}{ccc}
    \toprule
    Aux Loss & LayerScale & Acc. (\%) \\
    \midrule
    \ding{55} & \ding{55} & 73.9 \\
    \ding{51} & \ding{55} & 73.8 \\
    \ding{55} & \ding{51} & 74.0 \\
    \ding{51} & \ding{51} & \textbf{74.5} \\
    \bottomrule
  \end{tabular}
\end{minipage}
\hfill
\begin{minipage}[t]{0.48\textwidth}
  \centering
  \captionof{table}{Comparison of LayerScale variants and residual removal.}
  \label{tab:layerscale_variants}
  \begin{tabular}{lc}
    \toprule
    Configuration & Acc. (\%) \\
    \midrule
    Aux Loss only (baseline) & 73.8 \\
    Remove skip connection & 74.3 \\
    LayerScale (scalar) & 74.3 \\
    LayerScale (vector) & \textbf{74.5} \\
    LayerScale (scalar + linear) & 74.2 \\
    \bottomrule
  \end{tabular}
\end{minipage}

\vspace{-1.2em}
\subsection{Ablation Study}
\label{section:Ablation_Study}

To better understand the effectiveness of our design, we conduct ablation studies focusing on the key components of our method: the auxiliary loss $\mathcal{L}_\text{aux}$, the LayerScale mechanism, and their structural configurations. 
Unless otherwise specified, all ablation experiments are conducted by training Soft MoE from scratch on ImageNet-1K for 100 epochs, following the setup in Appendix~\ref{section:Experimental_Setups}.

\textbf{Effect of Components.}
We first examine the contribution of the auxiliary loss and LayerScale mechanism individually and in combination. 
As shown in Table~\ref{tab:ablation_components_of_method}, applying either individual component in isolation yields only marginal benefits, and in some cases, may even lead to a detrimental effect on performance.
When these elements are combined, there is a consistent and noticeable enhancement over the established baseline, demonstrating a clear advantage in performance.
Figure~\ref{fig:dispatch_ablation_visual} presents visualizations of the dispatch weight maps across various settings. 
The results vividly demonstrate that our proposed method effectively promotes a higher degree of semantic alignment in the activation of the experts. %
We refer to Appendix~\ref{app:multi-object} for examples with multiple objects.

\textbf{Auxiliary Loss Design.}
We conduct an in-depth study on the influence exerted by the weight $\lambda$ of the auxiliary loss, along with the exploration of its optimal application strategy. 
Figure~\ref{fig:effect_lambda} shows that $\lambda=0.01$ yields the best accuracy, and is thus used by default in our experiments. 
We further conduct tests on various variants,  including those that involve foreground v.s. background prior masks,  as well as different strategies for their placement.
Table~\ref{tab:aux_loss_variant}, where FG means foreground masks and BG means background masks, reveals that when the auxiliary loss is applied to the 8th layer with the utilization of a foreground mask, it consistently surpasses other configurations in terms of performance.
These results strongly imply that the quality of the spatial prior and the appropriate positioning play a crucial and indispensable role. 

\begin{minipage}[t]{0.48\textwidth}
  \centering
    \captionof{figure}{Effect of loss weight $\lambda$ on accuracy.}
  \includegraphics[width=\linewidth]{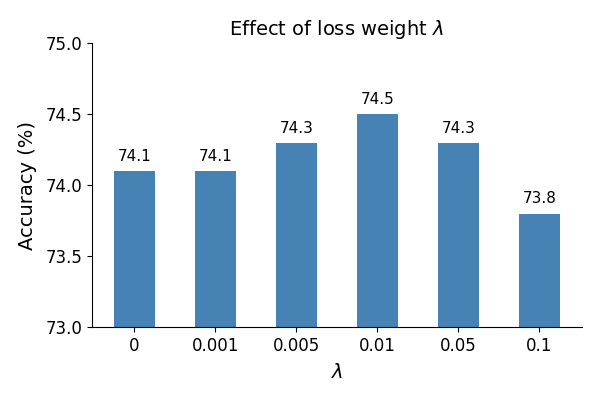}
  \label{fig:effect_lambda}
\end{minipage}%
\hfill
\begin{minipage}[t]{0.48\textwidth}
  \centering
  \captionof{table}{Effect of loss placement and mask type.}
  \label{tab:aux_loss_variant}
  \begin{tabular}{l l c}
    \toprule
    Placement & Mask Type & Acc. (\%) \\
    \midrule
    Baseline & None & 73.9 \\
    + 8th layer & FG & \textbf{74.5} \\
    + 8th layer & BG & 74.0 \\
    + 7th layer & FG & 73.5 \\
    + 7th layer & BG & 73.2 \\
    + 7\&8th layers & FG@7, BG@8 & 73.3 \\
    + 7\&8th layers & BG@7, FG@8 & 73.8 \\
    + 7\&8th layers & FG & 74.2 \\
    \bottomrule
  \end{tabular}
\end{minipage}

\vspace{-0.4em}
\textbf{LayerScale Variants.}
We conduct a comparison of several LayerScale configurations, encompassing a learnable scalar configuration, a vector-based configuration, and a configuration where a scalar is succeeded by a linear layer.
As shown in Table~\ref{tab:layerscale_variants}, 
channel-wise scaling via a learnable vector achieves the best performance, offering fine-grained modulation across dimensions. 
Scalar scaling shows performance on par with that of eliminating residual connections, which underscores its limited expressive capability. 
These observations suggest that fine-grained modulation provides more precise control and greater expressive capabilities, enabling more effective performance optimization.
The configuration that integrates a scalar with a linear layer exhibits the poorest performance.
We attribute this subpar outcome to the elevated structural complexity it introduces. This added complexity likely disrupts the smooth flow of residual information and gives rise to unstable optimization dynamics, ultimately hindering the model's effectiveness.
On large-scale datasets like ImageNet-1K, this also increases the risk of overfitting to non-generalizable routing patterns.

%% file: sections/5-2-analysis.tex
\paragraph{Foreground Consistency Dynamics.}
When training from scartch, we notice that the dispatch weights across various layers display clearly distinguishable consistent trends within the foreground area.
More specifically, within the baseline model, the 7th layer demonstrates a growing correspondence between high dispatch weights and foreground regions, whereas the 8th layer exhibits a declining tendency in this regard. 
Therefore, we design a variant that replaces the foreground mask in the auxiliary loss with a background mask to suppress foreground weights. 
As shown in Table~\ref{tab:aux_loss_variant}, it validates that the consistent utilization of foreground masks surpasses the employment of background masks in performance. 
This serves to confirm the practical advantage of foreground-aware expert routing, which is in line with the intuitive understanding that regions of semantic significance ought to attract more substantial attention.

\textbf{The placement of our method.} Regarding the placement of our method, we find that applying our method at the 7th layer leads to suboptimal performance. Visualization in Figure~\ref{fig:dispatch_ablation_visual}(c) shows that the increase in foreground weights in the 8th layer becomes insufficient. We suggest that inadequate optimization at this stage may introduce noise into the routing process and negatively affect performance.

\textbf{Shortcutting in Multi-Layer Loss Design.}
We conduct additional experiments by applying the auxiliary loss to the 7th and 8th layers concurrently.
In one setup, we applied foreground masks at the 7th layer and background masks at the 8th layer, matching the baseline's dispatch consistency trends. 
On the other hand, we reversed this, using background masks at the 7th and foreground masks at the 8th. Both setups caused performance drops.
We propose that conflicting supervision signals between layers may impede effective optimization.
We also test applying the auxiliary loss to both the 7th and 8th MoE layers. Intriguingly, this setup underperforms compared to applying the loss only to the 8th layer, even with what appears to be more robust supervision.
Figure~\ref{fig:dispatch_ablation_visual}(d) reveals that, in this case, the 8th layer's dispatch weights fail to fully concentrate on foreground areas. 
We hypothesize that the model prioritizes an easier optimization trajectory by overfitting to the 7th layer, which dominates loss reduction. 
This phenomenon is an analogous to a form of \textit{loss shortcutting}, in which  
earlier layers capture the bulk of the supervision signal. 
Consequently, this diminishes the impetus for deeper layers to adapt, leaving the more semantically crucial 8th layer insufficiently optimized.

These results altogether indicate 
two key insights for effective auxiliary supervision in Soft MoE: (1) it should be spatially congruent with significant foreground areas, and (2) %
it should be applied at the last expert-routing layer to guarantee ideal information dissemination and semantic emphasis.

%% file: sections/6-conclusion.tex
\section{Conclusion}
\label{section:Conclusion}
In this paper, we propose a novel approach to enhance the performance and interpretability of Soft MoE models by leveraging foreground-aware supervision. By incorporating a foreground-guided auxiliary loss, we guide the dispatch weights of the experts to focus on semantically meaningful foreground regions, improving the overall routing behavior. Additionally, we introduce a lightweight LayerScale mechanism to modulate residual connections, further stabilizing training and facilitating more effective expert specialization.
Through extensive experiments on ImageNet-1K and multiple smaller-scale classification benchmarks, we demonstrate that our method consistently outperforms baseline models and existing Soft MoE variants. Our approach not only improves performance but also leads to more interpretable expert routing mechanisms, providing valuable insights into the inner workings of MoE models. The visualizations confirm that our method aligns with the foreground semantics, even in complex multi-object scenes.

\textbf{Limitation.} A limitation of our method is that its performance relies on the quality of the foreground prior masks. Poor-quality masks may hinder the model's ability to effectively learn foreground features. Furthermore, our experiments focus primarily on image classification tasks, and we have not yet tested the method on other tasks such as object detection or semantic segmentation, which will be explored in future work. Nevertheless, our method shows consistent improvements in performance across datasets of various sizes, both during pretraining and fine-tuning, further validating its effectiveness.

%% file: sections/appendix.tex
\appendix

\section{Implementation of Aux Loss}
\label{section:implementation_of_aux_loss}
Algorithm~\ref{alg:aux_loss_combined} outlines the computation of our proposed foreground-aware auxiliary loss. The method first converts the dispatch weights into a binary mask by thresholding each token's weight against the sample-wise mean. This serves as an implicit indicator of expert attention. A prior foreground mask, extracted using an external segmentation method, is resized and binarized for alignment comparison.

To ensure meaningful supervision, we filter out samples with empty foreground regions. For valid samples, we compute the intersection and union between the dispatch mask and the foreground mask. A soft importance-weighted overlap score is then calculated, which reflects how much expert attention aligns with the foreground. The final loss is computed as the negative logarithm of this score, encouraging the model to focus on semantically important regions.

This loss operates in parallel with the main classification objective and is lightweight, fully differentiable, and applicable during pretraining or fine-tuning phases.
\begin{algorithm}[t]
\caption{Foreground-Aware Auxiliary Loss with Mask Validation}
\label{alg:aux_loss_combined}
\begin{algorithmic}[1]
\STATE \textbf{Description:} This algorithm computes an auxiliary loss that encourages alignment between 
high-dispatch-weight regions and semantic foreground masks. 
It uses mean-thresholded dispatch weights to generate a binary attention mask, 
compares it with a foreground prior, and computes a soft overlap ratio. 
Invalid samples (with empty masks) are filtered before aggregation.

\REQUIRE Dispatch weights $x \in \mathbb{R}^{B \times N}$, foreground masks $\mathcal{M} \in \mathbb{R}^{B \times 1 \times H \times W}$
\ENSURE Auxiliary loss $\mathcal{L}_{\text{aux}}$

\STATE $x \gets \text{mean}(x, \text{dim}=-1)$ \hfill // Collapse channel dim if needed
\STATE $l \gets \text{length}(x[0])$, $h, w \gets \sqrt{l}$ \hfill // Recover spatial map size
\STATE $\mathcal{M} \gets \text{Interpolate}(\mathcal{M}, \text{size}=(h,w))$ \hfill // Resize mask
\STATE $\mathcal{M} \gets (\mathcal{M} \ne 0)$ \hfill // Binarize foreground mask
\STATE $m \gets \text{mean}(x, \text{dim}=-1, \text{keepdim=True})$ \hfill // Token-wise mean per sample
\STATE $B_{\text{mask}} \gets (x \ge m)$ \hfill // Threshold weights to obtain binary attention mask

\STATE Initialize $\text{valid\_idx} \gets []$, $\mathcal{M}_{\text{valid}} \gets []$
\FOR{$b = 0$ to $B{-}1$}
    \IF{$\sum(\mathcal{M}[b]) \ne 0$}
        \STATE Append $b$ to $\text{valid\_idx}$ \hfill // Keep valid samples
        \STATE Append $\mathcal{M}[b]$ to $\mathcal{M}_{\text{valid}}$
    \ENDIF
\ENDFOR

\STATE $x \gets x[\text{valid\_idx}], \quad B_{\text{mask}} \gets B_{\text{mask}}[\text{valid\_idx}]$
\STATE $\text{inner} \gets B_{\text{mask}} \land \mathcal{M}_{\text{valid}}$ \hfill // Compute intersection
\STATE $\text{outer} \gets B_{\text{mask}} \lor \mathcal{M}_{\text{valid}}$ \hfill // Compute union

\STATE $p_{\text{num}} \gets \sum(x \cdot \text{inner}, \text{dim}=-1)$ \hfill // Weighted sum over intersection
\STATE $p_{\text{den}} \gets \sum(x \cdot \text{outer}, \text{dim}=-1)$ \hfill // Weighted sum over union
\STATE $p \gets p_{\text{num}} / (p_{\text{den}} + \epsilon)$ \hfill // Soft IoU-like score
\STATE $h \gets -\log(p)$ \hfill // Log loss to penalize misalignment

\STATE \textbf{return} $\mathcal{L}_{\text{aux}} \gets \text{mean}(h)$ \hfill // Mean across valid samples
\end{algorithmic}
\end{algorithm}

\section{Experimental Setups}
\label{section:Experimental_Setups}
To ensure reproducibility and facilitate further comparison, we summarize the key training configurations and hyperparameters used in our experiments. These settings are consistent across all evaluation scenarios and are detailed in Table~\ref{tab:basic_settings} and Table~\ref{tab:finetune_settings}.
\textbf{Dataset.} We conduct our experiments on 6 image classification datasets, including ImageNet-1K~\cite{russakovsky2015imagenet}, ImageNet-100~\cite{tian2020contrastive}, Stanford Cars~\cite{krause20133d}, and three datasets in DomainNet~\cite{peng2019moment} (\textit{i.e.}, Clipart, Painting and Sketch). 
As medium to large scale datasets, ImageNet-100 and ImageNet-1K are employed to assess the efficacy of our method when training from scratch. 
Meanwhile, Stanford Cars, Clipart, Painting, and Sketch represent small scale datasets commonly used in fine-tuning scenarios. 
These datasets allow us to thoroughly assess the adaptability and generalization of our method when transferring pre-trained features to downstream tasks characterized by a scarcity of data.
Through comprehensive evaluations across these two distinct settings, our objective is to vividly illustrate the broad applicability and remarkable robustness of our method, showcasing its efficacy in handling a wide spectrum of data scenarios.

\textbf{Model.} We validate our method by modifying the pytorch implementation of the Soft MoE model~\cite{fkodom_softmoe}.
More precisely, our model is composed of eight ViT~\cite{dosovitskiy2020image} blocks.
Notably, the MLP modules within the final two blocks are substituted with Soft MoE layers. Each of these Soft MoE layers houses 32 experts, and a single slot is allocated to each individual expert. 
This configuration follows the practice adopted in V-MoE~\cite{riquelme2021scaling}, where the authors utilized an 8-layer backbone architecture. 
They incorporated MoE modules within the final two layers of this backbone when performing ImageNet-1K classification tasks. 
This configuration has been shown to work well on ImageNet-1K and provides a lightweight yet effective setup for investigating our auxiliary loss design.

\textbf{Implementation.}
Since the authors of Soft MoE do not release pretrained checkpoints or evaluation results on the ImageNet-1K dataset, we reproduct the Soft MoE baseline using PyTorch with the Distributed Data Parallel (DDP)~\cite{li2020pytorch} framework. The training is conducted on 8 NVIDIA H800 GPUs. Our proposed method follows the same training strategy and environment for a fair comparison in subsequent experiments.
\begin{table}[h]
    \centering
    \caption{Summary of basic settings for pretrain experiments on ImageNet-1k and ImageNet-100. The configuration is consistent across baseline and our method.}
    \label{tab:basic_settings}
    \begin{tabular}{l l}
        \toprule
        \textbf{Hyperparameter} & \textbf{Value} \\
        \midrule
        \multicolumn{2}{l}{\textit{Training Procedure}} \\
        Optimizer               & AdamW \\
        Base Learning Rate      & 5e-4 \\
        Warmup Learning Rate    & 5e-7 \\
        Min Learning Rate       & 5e-6 \\
        Weight Decay            & 0.05 \\
        Learning Rate Schedule  & Cosine Decay \\
        Warmup Schedule         & Linear \\
        Warmup Epochs           & 30 \\
        Train Epochs            & 100 \\
        Batch Size (Global)     & 1024 \\
        Input Resolution        & $224 \times 224$ \\

        \midrule
        \multicolumn{2}{l}{\textit{MoE Architecture}} \\
        MoE Experts per Layer   & 32 \\
        MoE Layers              & Last 2 layers \\
        Slot per Expert         & 1 \\

        \midrule
        \multicolumn{2}{l}{\textit{Proposed Method Components}} \\
        Auxiliary Loss $\lambda$  & 0.01 \\
        LayerScale Init           & $\vec{0}$ \\

        \midrule
        \multicolumn{2}{l}{\textit{Data Augmentation}} \\
        Mixup                   & 0.8 \\
        Cutmix                  & 1.0 \\
        RandAugment             & (9, 0.5) \\
        \bottomrule
    \end{tabular}
\end{table}

\begin{table}[h]
    \centering
    \caption{Summary of basic settings for finetune experiments on. Other settings are consistent with Table~\ref{tab:basic_settings}}
    \label{tab:finetune_settings}
    \begin{tabular}{lcc}
        \toprule
       \textbf{Hyperparameter} & \textbf{Value}\\
        \midrule
        Base Learning Rate     & 2.5e-5\\
        Warmup Learning Rate     & 5e-7\\
        Min Learning Rate     & 1e-6\\
        Weight Decay      & 0.01 \\
        Batch Size        & 512               \\
        \bottomrule
    \end{tabular}
\end{table}

\section{Training efficiency}
\label{section:training_efficiency}
\begin{figure}[h]
    \centering
    \begin{subfigure}[t]{0.48\linewidth}
        \centering
        \includegraphics[width=\linewidth]{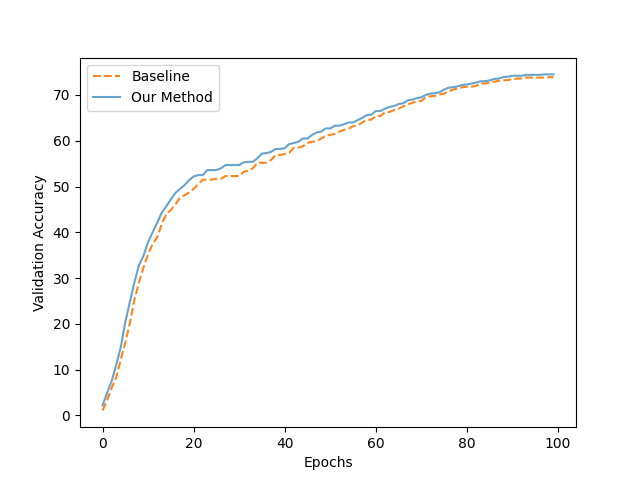}
        \caption{
            Validation accuracy curves.
            }
        \label{fig:val_acc}
    \end{subfigure}
    \begin{subfigure}[t]{0.48\linewidth}
        \centering
        \includegraphics[width=\linewidth]{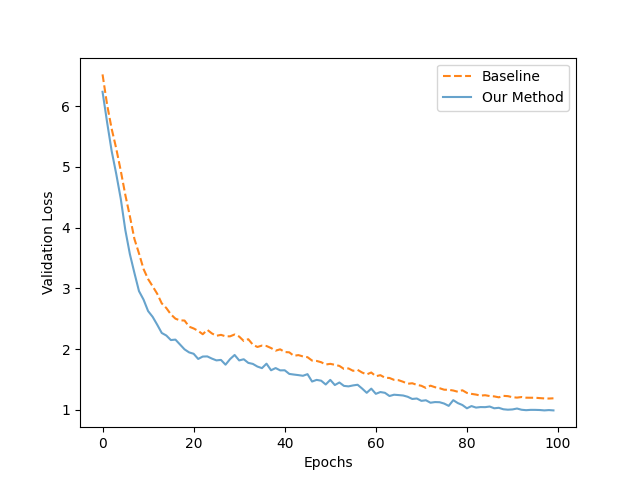}
        \caption{
            Validation loss curves.
            }
        \label{fig:val_loss}
    \end{subfigure}
    \caption{
    (a) Validation accuracy curves during training. 
            The model trained with our proposed routing loss (blue solid line) converges faster and consistently outperforms the baseline model (orange dotted line) across all epochs. 
            This demonstrates both improved convergence efficiency and better final performance.
    (b)Validation loss curves over 100 epochs. 
            Our method (blue solid line) leads to a more stable and faster reduction in training loss compared to the baseline (orange dotted line), suggesting improved optimization dynamics and training efficiency.
    }
    \label{fig:validation_vis}
\end{figure}

As shown in Figure~\ref{fig:val_acc}, the model trained with our auxiliary routing loss achieves faster convergence in the early stage and consistently outperforms the baseline throughout training. Notably, it reaches 60\% accuracy about 5 epochs earlier than the original model.
We further analyze the evolution of validation loss over epochs. As shown in Figure~\ref{fig:val_loss}, the model trained with the proposed auxiliary routing loss exhibits a faster reduction compared to the baseline. This suggests that our approach facilitates more efficient optimization and accelerates convergence during the early training phase.

Table~\ref{tab:early_acc} presents the average validation accuracy during the early stages of training. Our method consistently outperforms the baseline within the first 20 and 30 epochs, indicating more efficient learning in the initial optimization phase. This suggests that the auxiliary loss facilitates better dispatch weights from the beginning of training, enabling the model to extract discriminative features more rapidly.

\begin{table}[h]
\centering
\caption{Convergence speed and early-stage accuracy comparison.}
\label{tab:converge_combined}
\begin{subtable}[t]{0.44\linewidth}
    \centering
    \caption{Epochs to reach threshold validation accuracy}
    \label{tab:converge_speed}
    \begin{tabular}{lccc}
    \toprule
    \textbf{Method} & 50\% & 60\% & 73\% \\
    \midrule
    Baseline        & 21   & 48   & 78   \\
    Ours Method     & \textbf{18} & \textbf{43} & \textbf{73} \\
    \bottomrule
    \end{tabular}
\end{subtable}
\hfill
\begin{subtable}[t]{0.55\linewidth}
    \centering
    \caption{Average validation accuracy over early epochs}
    \label{tab:early_acc}
    \begin{tabular}{lcc}
    \toprule
    \textbf{Method} & Epoch 1--20 & Epoch 1--30 \\
    \midrule
    Baseline        & 29.3\%     & 36.7\% \\
    Ours (Routing Loss) & \textbf{32.0\%} & \textbf{39.2\%} \\
    \bottomrule
    \end{tabular}
\end{subtable}
\end{table}

\section{More Visualizations}
\label{section:more_visualizations}
\begin{figure}[ht]
  \centering
  \begin{subfigure}[t]{0.22\textwidth}
    \includegraphics[width=\textwidth]{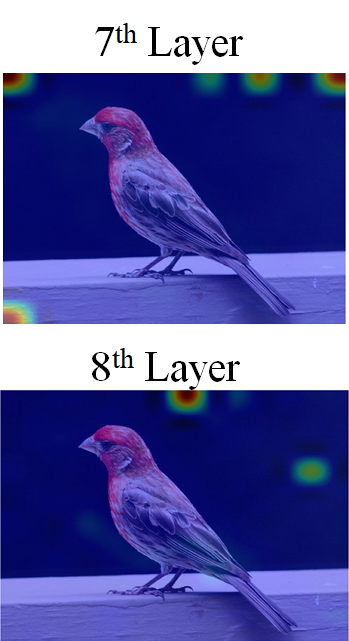}
    \caption{Background Loss on 7th layer}
  \end{subfigure}
  \hspace{0.01\textwidth}
  \begin{subfigure}[t]{0.22\textwidth}
    \includegraphics[width=\textwidth]{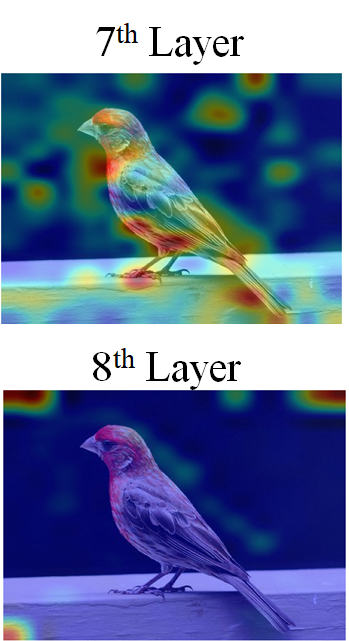}
    \caption{Background Loss on 8th layer}
  \end{subfigure}
  \hspace{0.01\textwidth}
  \begin{subfigure}[t]{0.22\textwidth}
    \includegraphics[width=\textwidth]{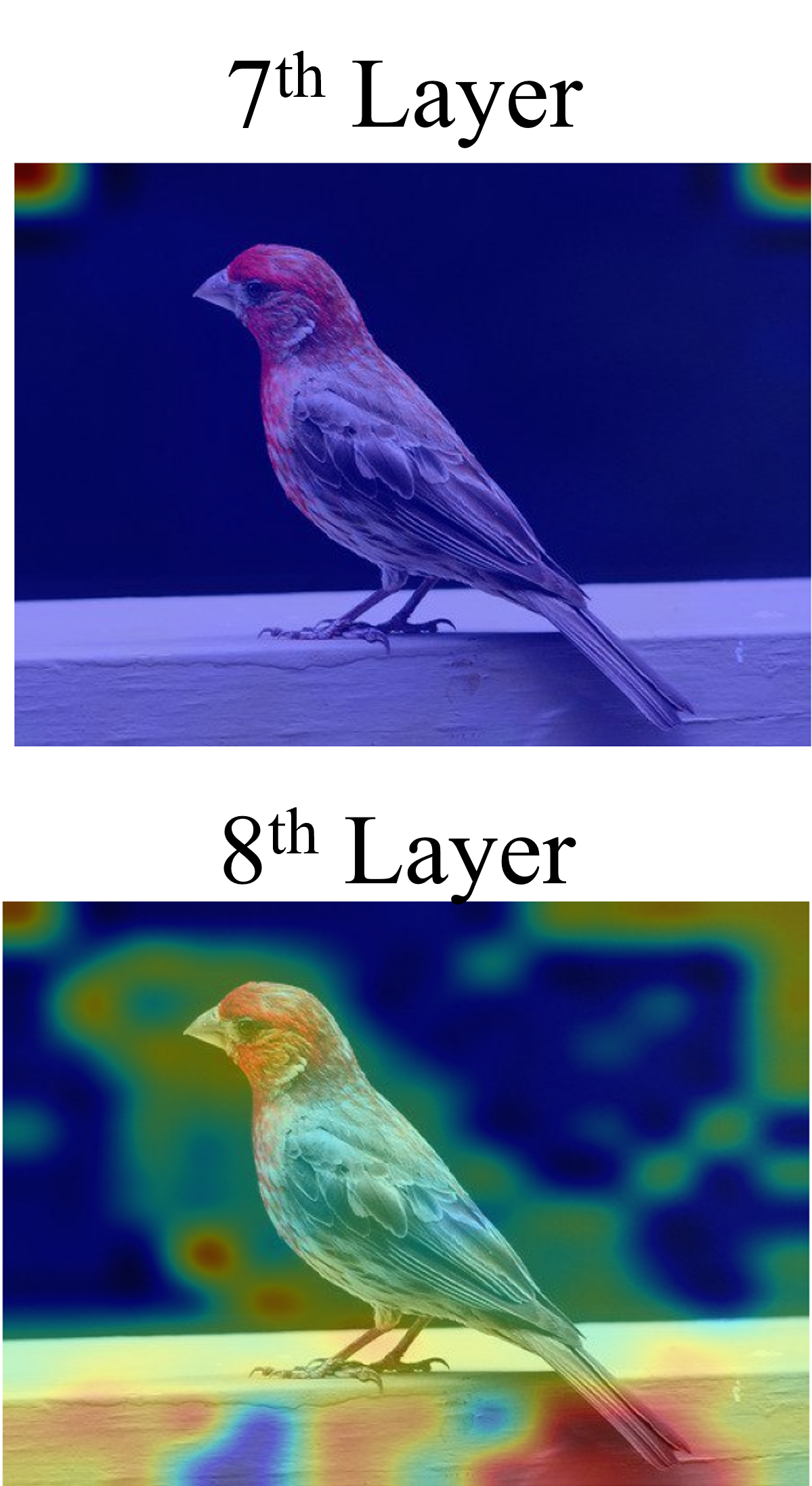}
    \caption{Background Loss on 7th layer and Foreground Loss on 8th layer}
  \end{subfigure}
    \hspace{0.01\textwidth}
  \begin{subfigure}[t]{0.22\textwidth}
    \includegraphics[width=\textwidth]{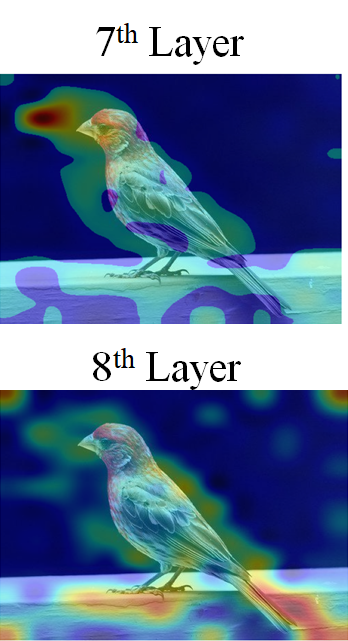}
    \caption{Foreground Loss on 7th layer and Background Loss on 8th layer}
  \end{subfigure}
  \caption{
    Visualization of dispatch weights under different auxiliary loss, corresponding to the settings reported in Table~\ref{tab:aux_loss_variant}: (a) applying background loss to the 7th layer, (b) applying background loss to the 8th layer, (v) using foreground loss on the 7th layer and background loss on the 8th layer, and (d) the reverse setup of (3).
    }
\label{fig:vis_extra_loss_variants}
\end{figure}

In this section, we provide additional visualizations that were not included in the main experimental results. Figure~\ref{fig:vis_extra_loss_variants} illustrates the dispatch weight maps of models trained from scratch on ImageNet-1K under four different auxiliary loss configurations.
\begin{figure}[ht]
  \centering
  \begin{subfigure}[t]{0.22\textwidth}
    \includegraphics[width=\textwidth]{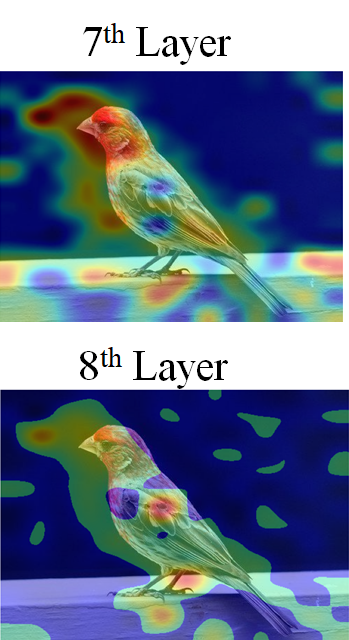}
    \caption{LayerScale based on scale}
  \end{subfigure}
  \hspace{0.01\textwidth}
  \begin{subfigure}[t]{0.22\textwidth}
    \includegraphics[width=\textwidth]{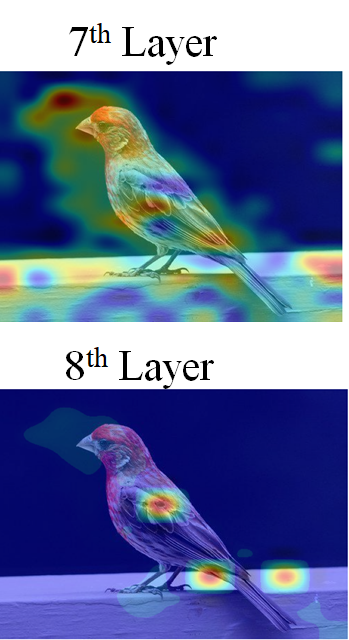}
    \caption{LayerScale based on linear layer}
  \end{subfigure}
   \hspace{0.01\textwidth}
  \begin{subfigure}[t]{0.22\textwidth}
    \includegraphics[width=\textwidth]{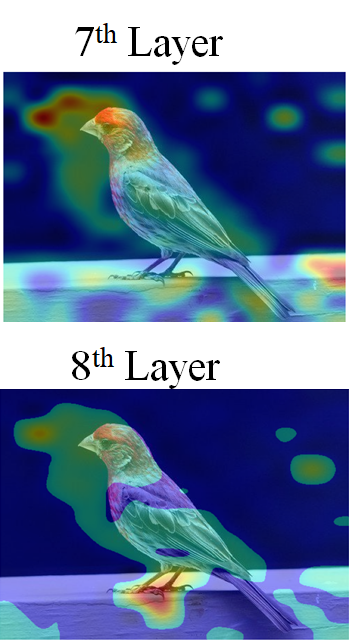}
    \caption{Removal of skip connection}
  \end{subfigure}
\caption{
        Visualization of dispatch weights under different LayerScale configurations in the last Soft MoE layer. 
        (a) Scaling with a learnable scalar shows relatively focused but less adaptive attention. 
        (b) Using a learnable linear transformation leads to dispersed and weak expert activation, indicating unstable routing. 
        (c) Removing the skip connection entirely yields better semantic focus than (b), but still underperforms vector-based LayerScale (not shown). 
        }

\label{fig:vis_extra_layerscale_variants}
\end{figure}
From the visualizations, it is evident that applying background loss successfully drives expert attention away from the semantic foreground. However, this also leads to more dispersed attention and even negatively impacts routing behavior in earlier layers such as the 6th layer. These findings further support our conclusion in the main paper: semantic foreground regions are inherently more informative than background regions. Attempting to guide expert routing toward background areas via auxiliary loss results in unstable learning dynamics and degrades overall performance.

We also explore combining different losses across layers in figure~\ref{fig:vis_extra_layerscale_variants}. The results reveal conflicting effects between foreground and background loss, where the foreground loss tends to dominate. However, this dominance creates a mismatch in supervision across layers, making it harder for the auxiliary loss to consistently guide the model toward learning semantically meaningful foreground regions.

\begin{figure}  
    \centering
    \includegraphics[width=\textwidth]{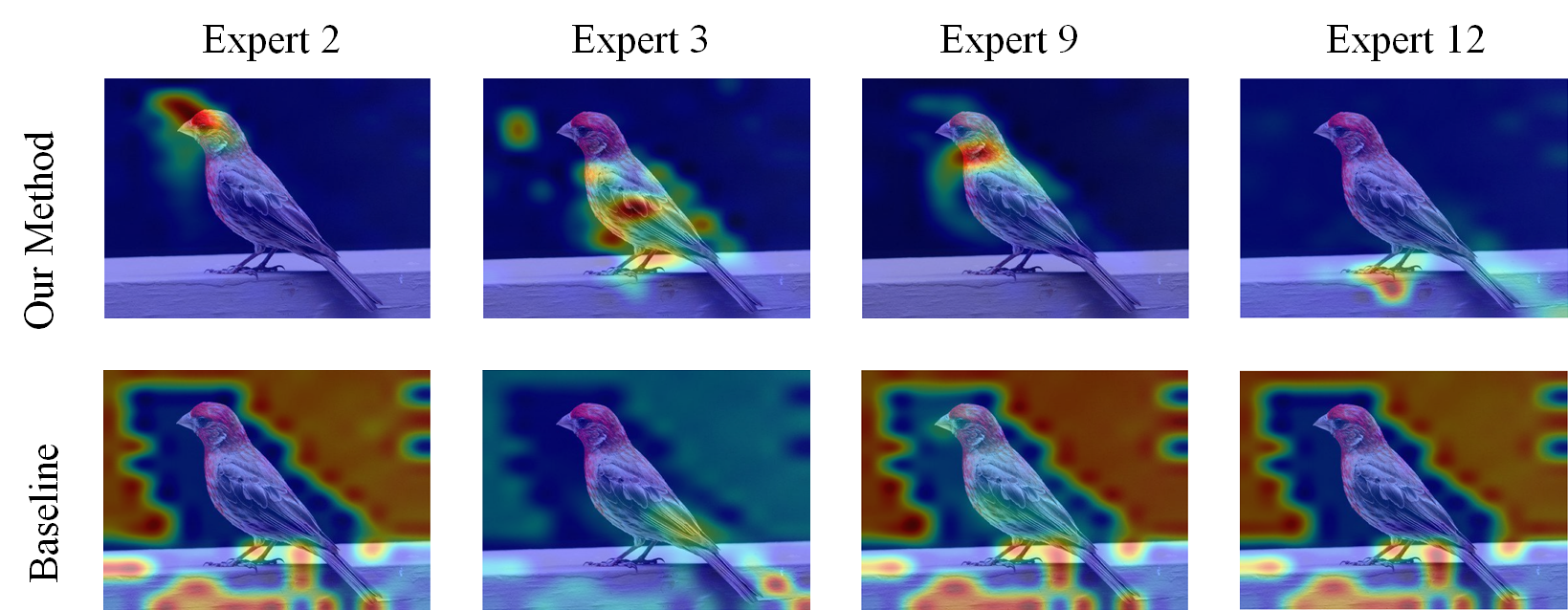} 
    \caption{
        Visualization of individual expert activations from the 8th Soft MoE layer. 
        }
\label{fig:aux_layerscale_expert_visual}
\end{figure}

In addition, we visualize different structural variants of the LayerScale mechanism. Among them, the "scale + linear" configuration performs the worst, as the resulting dispatch weight maps in the 8th layer show unfocused attention with weak activation magnitudes. While both the scalar-only variant and the skip-connection removal exhibit better foreground alignment, they still underperform compared to the vector-based LayerScale. This suggests that channel-wise modulation via a learnable vector provides more precise control over expert activation and facilitates stronger spatial alignment with semantic foreground regions.
Beyond analyzing the overall behavior of the expert layer, we further examine the activation patterns of individual experts. As shown in Figure~\ref{fig:aux_layerscale_expert_visual}, the baseline model exhibits limited diversity among experts, with most of them focusing predominantly on background regions. In contrast, our method enables each expert to attend to different parts of the semantic foreground, promoting better specialization and collaboration across experts.
It is worth noting that our model is trained on the ImageNet-1K dataset, which is substantially smaller than the JFT-4B dataset used in the original Soft MoE paper. This reduced data scale may hinder the emergence of well-separated expert dispatch strategies. However, our proposed auxiliary loss effectively mitigates this issue by guiding expert routing toward more meaningful regions, leading to improved training stability and final performance.

\section{Broader Impact}
\label{section:broader_impact}
Our work proposes a foreground-aware routing strategy for Soft Mixture-of-Experts (MoE) models in vision tasks, aiming to improve expert activation alignment with semantically meaningful regions. This approach enhances both the interpretability and efficiency of large-scale sparse models, which may have broader implications in domains where model transparency and data efficiency are critical.

By guiding expert attention toward semantically relevant foreground areas, our method can potentially reduce reliance on extremely large datasets for effective MoE training. This makes the technique more accessible to research communities with limited computational resources, and may enable the deployment of efficient, high-performing models in edge scenarios or privacy-sensitive domains such as medical imaging or autonomous driving.

However, as with any method that incorporates prior information (e.g., segmentation masks), care must be taken to avoid propagating dataset-specific biases or errors in foreground definition. If the foreground masks reflect biased annotations or semantic misinterpretation, this may unintentionally guide experts toward irrelevant or misleading regions, potentially degrading fairness or generalization.

In future extensions of this work, we recommend further exploration of domain-adaptive or unsupervised foreground estimation techniques to improve robustness. Moreover, given the increasing popularity of MoE models in foundation model infrastructure, ensuring that expert routing remains stable, efficient, and semantically grounded is essential for building reliable large-scale AI systems.

\section{Multi-object Scene}
\label{app:multi-object}
\begin{figure}[h]
    \centering
    \begin{subfigure}[t]{0.23\linewidth}
        \centering
        \includegraphics[width=\linewidth]{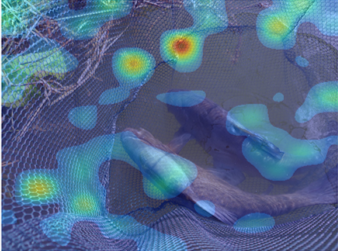}
        \caption{Baseline (two fish)}
        \label{fig:multi_object_baseline}
    \end{subfigure}
    \begin{subfigure}[t]{0.23\linewidth}
        \centering
        \includegraphics[width=\linewidth]{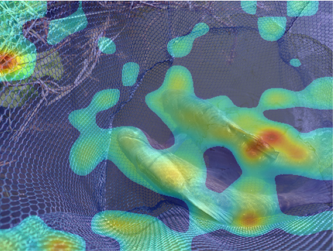}
        \caption{Ours (two fish)}
        \label{fig:multi_object_ours}
    \end{subfigure}
    \begin{subfigure}[t]{0.23\linewidth}
        \centering
        \includegraphics[width=\linewidth]{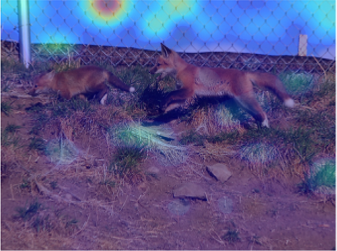}
        \caption{Baseline (two foxes)}
        \label{fig:multi_object2_baseline}
    \end{subfigure}
    \begin{subfigure}[t]{0.23\linewidth}
        \centering
        \includegraphics[width=\linewidth]{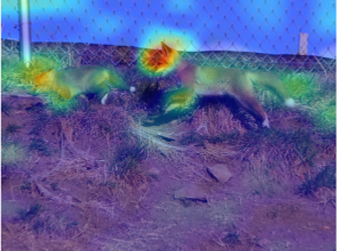}
        \caption{Ours (two foxes)}
        \label{fig:multi_object2_ours}
    \end{subfigure}

    \caption{
    Dispatch weight visualizations on multi-object images. 
    (a) and (b) compare the baseline and our method on an image containing two fish. 
    (c) and (d) show a similar comparison on an image with two foxes. 
    Our method demonstrates improved alignment with semantic foreground regions, even in scenes with multiple entities.
    }
    \label{fig:multi_object_visual}
\end{figure}

To further demonstrate the spatial effectiveness of our auxiliary loss, we visualize the dispatch weight maps on images containing multiple foreground objects. As shown in Figure~\ref{fig:multi_object_visual}, our method successfully focuses on semantically meaningful regions that correspond to the primary objects with in a scene, regardless of the presence of multiple entities. 
This demonstrates that the routing behavior acquired through learning is not only sensitive to foreground elements but also resilient in the face of intricate spatial arrangements.

%% file: arxiv.bbl
\begin{thebibliography}{10}

\bibitem{carion2020end}
Nicolas Carion, Francisco Massa, Gabriel Synnaeve, Nicolas Usunier, Alexander Kirillov, and Sergey Zagoruyko.
\newblock End-to-end object detection with transformers.
\newblock In {\em European Conference on Computer Vision (ECCV)}, 2020.

\bibitem{caron2021emerging}
Mathilde Caron, Hugo Touvron, Ishan Misra, Herv\'e J\'egou, Julien Mairal, Piotr Bojanowski, and Armand Joulin.
\newblock Emerging properties in self-supervised vision transformers.
\newblock In {\em Proceedings of the IEEE/CVF International Conference on Computer Vision (ICCV)}, 2021.

\bibitem{dosovitskiy2020image}
Alexey Dosovitskiy, Lucas Beyer, Alexander Kolesnikov, Dirk Weissenborn, Xiaohua Zhai, Thomas Unterthiner, Mostafa Dehghani, Matthias Minderer, Georg Heigold, Sylvain Gelly, et~al.
\newblock An image is worth 16x16 words: Transformers for image recognition at scale.
\newblock {\em arXiv preprint arXiv:2010.11929}, 2020.

\bibitem{fedus2022switch}
William Fedus, Barret Zoph, and Noam Shazeer.
\newblock Switch transformers: Scaling to trillion parameter models with simple and efficient sparsity.
\newblock {\em Journal of Machine Learning Research}, 23(120):1--39, 2022.

\bibitem{he2016deep}
Kaiming He, Xiangyu Zhang, Shaoqing Ren, and Jian Sun.
\newblock Deep residual learning for image recognition.
\newblock In {\em Proceedings of the IEEE Conference on Computer Vision and Pattern Recognition (CVPR)}, 2016.

\bibitem{hwang2023tutel}
Changho Hwang, Wei Cui, Yifan Xiong, Ziyue Yang, Ze~Liu, Han Hu, Zilong Wang, Rafael Salas, Jithin Jose, Prabhat Ram, et~al.
\newblock Tutel: Adaptive mixture-of-experts at scale.
\newblock {\em Proceedings of Machine Learning and Systems}, 5:269--287, 2023.

\bibitem{jiang2024mixtral}
Albert~Q Jiang, Alexandre Sablayrolles, Antoine Roux, Arthur Mensch, Blanche Savary, Chris Bamford, Devendra~Singh Chaplot, Diego de~las Casas, Emma~Bou Hanna, Florian Bressand, et~al.
\newblock Mixtral of experts.
\newblock {\em arXiv preprint arXiv:2401.04088}, 2024.

\bibitem{kaplan2020scaling}
Jared Kaplan, Sam McCandlish, Tom Henighan, Tom~B Brown, Benjamin Chess, Rewon Child, Scott Gray, Alec Radford, Jeffrey Wu, and Dario Amodei.
\newblock Scaling laws for neural language models.
\newblock {\em arXiv preprint arXiv:2001.08361}, 2020.

\bibitem{kirillov2023segment}
Alexander Kirillov, Eric Mintun, Nikhila Ravi, Hanzi Mao, Chloe Rolland, Laura Gustafson, Tete Xiao, Spencer Whitehead, Alexander~C Berg, Wan-Yen Lo, et~al.
\newblock Segment anything.
\newblock In {\em Proceedings of the IEEE/CVF International Conference on Computer Vision (ICCV)}, 2023.

\bibitem{fkodom_softmoe}
Forest Kodom.
\newblock Soft mixture of experts - official implementation.
\newblock \url{https://github.com/fkodom/soft-mixture-of-experts}, 2023.
\newblock Accessed: 2025-05-16.

\bibitem{krause20133d}
Jonathan Krause, Michael Stark, Jia Deng, and Li~Fei-Fei.
\newblock 3d object representations for fine-grained categorization.
\newblock In {\em Proceedings of the IEEE International Conference on Computer Vision Workshops}, 2013.

\bibitem{krizhevsky2012imagenet}
Alex Krizhevsky, Ilya Sutskever, and Geoffrey~E Hinton.
\newblock Imagenet classification with deep convolutional neural networks.
\newblock {\em Advances in neural information processing systems}, 25, 2012.

\bibitem{lepikhin2020gshard}
Dmitry Lepikhin, HyoukJoong Lee, Yuanzhong Xu, Dehao Chen, Orhan Firat, Yanping Huang, Maxim Krikun, Noam Shazeer, and Zhifeng Chen.
\newblock Gshard: Scaling giant models with conditional computation and automatic sharding.
\newblock {\em arXiv preprint arXiv:2006.16668}, 2020.

\bibitem{lewis2021base}
Mike Lewis, Shruti Bhosale, Tim Dettmers, Naman Goyal, and Luke Zettlemoyer.
\newblock Base layers: Simplifying training of large, sparse models.
\newblock In {\em International Conference on Machine Learning (ICML)}, 2021.

\bibitem{li2022grounded}
Liunian~Harold Li, Pengchuan Zhang, Haotian Zhang, Jianwei Yang, Chunyuan Li, Yiwu Zhong, Lijuan Wang, Lu~Yuan, Lei Zhang, Jenq-Neng Hwang, et~al.
\newblock Grounded language-image pre-training.
\newblock In {\em Proceedings of the IEEE/CVF Conference on Computer Vision and Pattern Recognition (CVPR)}, 2022.

\bibitem{li2020pytorch}
Shen Li, Yanli Zhao, Rohan Varma, Omkar Salpekar, Pieter Noordhuis, Teng Li, Adam Paszke, Jeff Smith, Brian Vaughan, Pritam Damania, et~al.
\newblock Pytorch distributed: Experiences on accelerating data parallel training.
\newblock {\em arXiv preprint arXiv:2006.15704}, 2020.

\bibitem{liu2024deepseekv2}
Aixin Liu, Bei Feng, Bin Wang, Bingxuan Wang, Bo~Liu, Chenggang Zhao, Chengqi Dengr, Chong Ruan, Damai Dai, Daya Guo, et~al.
\newblock Deepseek-v2: A strong, economical, and efficient mixture-of-experts language model.
\newblock {\em arXiv preprint arXiv:2405.04434}, 2024.

\bibitem{liu2024deepseekv3}
Aixin Liu, Bei Feng, Bing Xue, Bingxuan Wang, Bochao Wu, Chengda Lu, Chenggang Zhao, Chengqi Deng, Chenyu Zhang, Chong Ruan, et~al.
\newblock Deepseek-v3 technical report.
\newblock {\em arXiv preprint arXiv:2412.19437}, 2024.

\bibitem{liu2024grounding}
Shilong Liu, Zhaoyang Zeng, Tianhe Ren, Feng Li, Hao Zhang, Jie Yang, Qing Jiang, Chunyuan Li, Jianwei Yang, Hang Su, et~al.
\newblock Grounding dino: Marrying dino with grounded pre-training for open-set object detection.
\newblock In {\em European Conference on Computer Vision (ECCV)}, 2024.

\bibitem{liu2021swin}
Ze~Liu, Yutong Lin, Yue Cao, Han Hu, Yixuan Wei, Zheng Zhang, Stephen Lin, and Baining Guo.
\newblock Swin transformer: Hierarchical vision transformer using shifted windows.
\newblock In {\em Proceedings of the IEEE/CVF International Conference on Computer Vision (ICCV)}, 2021.

\bibitem{liu2022convnet}
Zhuang Liu, Hanzi Mao, Chao-Yuan Wu, Christoph Feichtenhofer, Trevor Darrell, and Saining Xie.
\newblock A convnet for the 2020s.
\newblock In {\em Proceedings of the IEEE/CVF Conference on Computer Vision and Pattern Recognition (CVPR)}, 2022.

\bibitem{locatello2020object}
Francesco Locatello, Dirk Weissenborn, Thomas Unterthiner, Aravindh Mahendran, Georg Heigold, Jakob Uszkoreit, Alexey Dosovitskiy, and Thomas Kipf.
\newblock Object-centric learning with slot attention.
\newblock {\em Advances in Neural Information Processing Systems (NeurIPS)}, 2020.

\bibitem{medeiros2023langsam}
Luca Medeiros.
\newblock Language segment anything.
\newblock \url{https://github.com/luca-medeiros/lang-segment-anything}, 2023.
\newblock Accessed: 2025-05-16.

\bibitem{muqeeth2023soft}
Mohammed Muqeeth, Haokun Liu, and Colin Raffel.
\newblock Soft merging of experts with adaptive routing.
\newblock {\em arXiv preprint arXiv:2306.03745}, 2023.

\bibitem{naseer2021intriguing}
Muhammad~Muzammal Naseer, Kanchana Ranasinghe, Salman~H Khan, Munawar Hayat, Fahad Shahbaz~Khan, and Ming-Hsuan Yang.
\newblock Intriguing properties of vision transformers.
\newblock {\em Advances in Neural Information Processing Systems (NeurIPS)}, 2021.

\bibitem{oquab2023dinov2}
Maxime Oquab, Timoth{\'e}e Darcet, Th{\'e}o Moutakanni, Huy Vo, Marc Szafraniec, Vasil Khalidov, Pierre Fernandez, Daniel Haziza, Francisco Massa, Alaaeldin El-Nouby, et~al.
\newblock Dinov2: Learning robust visual features without supervision.
\newblock {\em arXiv preprint arXiv:2304.07193}, 2023.

\bibitem{peng2019moment}
Xingchao Peng, Qinxun Bai, Xide Xia, Zijun Huang, Kate Saenko, and Bo~Wang.
\newblock Moment matching for multi-source domain adaptation.
\newblock In {\em Proceedings of the IEEE/CVF International Conference on Computer Vision (ICCV)}, 2019.

\bibitem{peruzzo2024spatial}
Elia Peruzzo, Enver Sangineto, Yahui Liu, Marco De~Nadai, Wei Bi, Bruno Lepri, and Nicu Sebe.
\newblock Spatial entropy as an inductive bias for vision transformers.
\newblock {\em Machine Learning}, 113(9):6945--6975, 2024.

\bibitem{puigcerver2023sparse}
Joan Puigcerver, Carlos Riquelme, Basil Mustafa, and Neil Houlsby.
\newblock From sparse to soft mixtures of experts.
\newblock {\em arXiv preprint arXiv:2308.00951}, 2023.

\bibitem{riquelme2021scaling}
Carlos Riquelme, Joan Puigcerver, Basil Mustafa, Maxim Neumann, Rodolphe Jenatton, Andr{\'e} Susano~Pinto, Daniel Keysers, and Neil Houlsby.
\newblock Scaling vision with sparse mixture of experts.
\newblock {\em Advances in Neural Information Processing Systems (NeurIPS)}, 2021.

\bibitem{roller2021hash}
Stephen Roller, Sainbayar Sukhbaatar, Jason Weston, et~al.
\newblock Hash layers for large sparse models.
\newblock {\em Advances in Neural Information Processing Systems (NeurIPS)}, 2021.

\bibitem{russakovsky2015imagenet}
Olga Russakovsky, Jia Deng, Hao Su, Jonathan Krause, Sanjeev Satheesh, Sean Ma, Zhiheng Huang, Andrej Karpathy, Aditya Khosla, Michael Bernstein, et~al.
\newblock Imagenet large scale visual recognition challenge.
\newblock {\em International Journal of Computer Vision (IJCV)}, 115:211--252, 2015.

\bibitem{shazeer2017outrageously}
Noam Shazeer, Azalia Mirhoseini, Krzysztof Maziarz, Andy Davis, Quoc Le, Geoffrey Hinton, and Jeff Dean.
\newblock Outrageously large neural networks: The sparsely-gated mixture-of-experts layer.
\newblock {\em arXiv preprint arXiv:1701.06538}, 2017.

\bibitem{tian2020contrastive}
Yonglong Tian, Dilip Krishnan, and Phillip Isola.
\newblock Contrastive multiview coding.
\newblock In {\em European Conference Computer Vision (ECCV)}, 2020.

\bibitem{touvron2021going}
Hugo Touvron, Matthieu Cord, Alexandre Sablayrolles, Gabriel Synnaeve, and Herv{\'e} J{\'e}gou.
\newblock Going deeper with image transformers.
\newblock In {\em Proceedings of the IEEE/CVF International Conference on Computer Vision (ICCV)}, 2021.

\bibitem{vaswani2017attention}
Ashish Vaswani, Noam Shazeer, Niki Parmar, Jakob Uszkoreit, Llion Jones, Aidan~N Gomez, {\L}ukasz Kaiser, and Illia Polosukhin.
\newblock Attention is all you need.
\newblock {\em Advances in neural information processing systems}, 30, 2017.

\bibitem{wang2023tokencut}
Yangtao Wang, Xi~Shen, Yuan Yuan, Yuming Du, Maomao Li, Shell~Xu Hu, James~L Crowley, and Dominique Vaufreydaz.
\newblock Tokencut: Segmenting objects in images and videos with self-supervised transformer and normalized cut.
\newblock {\em IEEE Transactions on Pattern Analysis and Machine Intelligence (TPAMI)}, 45(12):15790--15801, 2023.

\bibitem{wu2024mixture}
Xun Wu, Shaohan Huang, and Furu Wei.
\newblock Mixture of lora experts.
\newblock {\em arXiv preprint arXiv:2404.13628}, 2024.

\bibitem{zhai2022scaling}
Xiaohua Zhai, Alexander Kolesnikov, Neil Houlsby, and Lucas Beyer.
\newblock Scaling vision transformers.
\newblock In {\em Proceedings of the IEEE/CVF Conference on Computer Vision and Pattern Recognition (CVPR)}, 2022.

\bibitem{zhong2024lory}
Zexuan Zhong, Mengzhou Xia, Danqi Chen, and Mike Lewis.
\newblock Lory: Fully differentiable mixture-of-experts for autoregressive language model pre-training.
\newblock {\em arXiv preprint arXiv:2405.03133}, 2024.

\bibitem{zhou2021ibot}
Jinghao Zhou, Chen Wei, Huiyu Wang, Wei Shen, Cihang Xie, Alan Yuille, and Tao Kong.
\newblock ibot: Image bert pre-training with online tokenizer.
\newblock {\em arXiv preprint arXiv:2111.07832}, 2021.

\bibitem{zhou2022mixture}
Yanqi Zhou, Tao Lei, Hanxiao Liu, Nan Du, Yanping Huang, Vincent Zhao, Andrew~M Dai, Quoc~V Le, James Laudon, et~al.
\newblock Mixture-of-experts with expert choice routing.
\newblock {\em Advances in Neural Information Processing Systems (NeurIPS)}, 2022.

\end{thebibliography}
